# Long-VLA: Unleashing Long-Horizon Capability of Vision Language Action Model for Robot Manipulation

**Yiguo Fan**[1*]   **Pengxiang Ding**[1,2*†]   **Shuanghao Bai**[3*]   **Xinyang Tong**[1*]   **Yuyang Zhu**[5]
**Hongchao Lu**[1]   **Fengqi Dai**[1]   **Wei Zhao**[1]   **Yang Liu**[1]   **Siteng Huang**[1]   **Zhaoxin Fan**[4]
**Badong Chen**[3✉]   **Donglin Wang**[1✉]

[1]Westlake University   [2]Zhejiang University   [3]Xi'an Jiaotong University
[4]Beijing Advanced Innovation Center for Future Blockchain and Privacy Computing   [5]University of Electronic Science and Technology of China

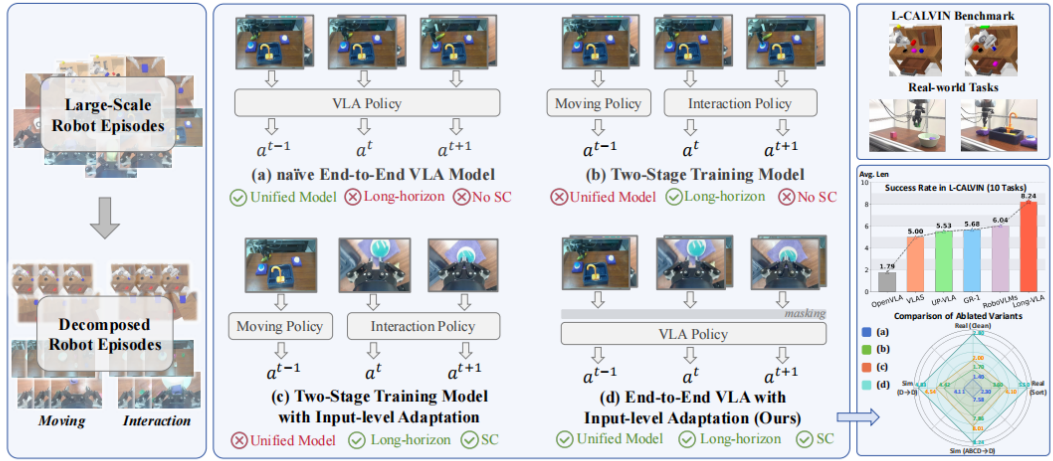

Figure 1: In contrast to previous methods that (a) adopt a unified model but are limited to short-horizon tasks and fail to address skill chaining (SC) [1], (b) decompose long-horizon tasks into moving and interaction stages using two separate models, reducing learning complexity but still unable to solve SC [2], and (c) further introduce adaptive input strategies on top of decomposition to address SC [3], (d) our Long-VLA is a unified model tailored for long-horizon tasks, and further incorporates input-level adaptation via masking to effectively solve SC. Long-VLA surpasses prior state-of-the-art methods with consistent performance and strong robustness.

**Abstract:** Vision-Language-Action (VLA) models have become a cornerstone in robotic policy learning, leveraging large-scale multimodal data for robust and scalable control. However, existing VLA frameworks primarily address short-horizon tasks, and their effectiveness on long-horizon, multi-step robotic manipulation remains limited due to challenges in skill chaining and subtask dependencies. In this work, we introduce Long-VLA, the first end-to-end VLA model specifically designed for long-horizon robotic tasks. Our approach features a novel phase-aware input masking strategy that adaptively segments each subtask into moving and interaction phases, enabling the model to focus on phase-relevant sensory cues and enhancing subtask compatibility. This unified strategy preserves the scalability and data efficiency of VLA training, and our architecture-agnostic module can be seamlessly integrated into existing VLA models. We further propose the L-CALVIN benchmark to systematically evaluate long-horizon manipulation.

\* Equal Contribution   † Project Leader   ✉ Equal Corresponding Author

Extensive experiments on both simulated and real-world tasks demonstrate that Long-VLA significantly outperforms prior state-of-the-art methods, establishing a new baseline for long-horizon control. Our project page is this https URL.

## 1 Introduction

Vision-Language-Action (VLA) models [4, 5, 6, 7, 8, 9, 10, 11, 12, 13, 14, 15] have achieved widespread adoption in robotic control, owing to their ability to leverage large-scale robot data for robust policy learning. However, most existing VLA frameworks are tailored for short-horizon tasks, leaving the challenge of long-horizon task execution largely unresolved.

Recent mainstream approaches [16, 17] address long-horizon tasks by decomposing them into sequences of subtasks, each managed by a separate local policy. Although this strategy reduces the complexity of learning individual behaviors, it does not adequately account for the transitions and dependencies between subtasks, commonly referred to as the skill chaining problem [18, 19]. In practice, this often leads to dynamic coupling and error propagation across subtask boundaries, which significantly degrades overall task performance.

Although various approaches [20, 21, 22, 19, 3, 23, 24] from related fields, such as online adaptive optimization [21] and decoupling motion planning and execution using different input modalities [3], have been proposed to address skill chaining, they are often incompatible with the end-to-end, scalable training paradigm central to VLA models. For example, reward-driven online methods [21] are difficult to reconcile with the large-scale offline training regime typical of VLA models, while modular architectures [3] hinder joint end-to-end learning and contradict the unified, data-driven philosophy of VLA frameworks. Therefore, solving the skill chaining problem in long-horizon tasks while preserving the scalability and data efficiency of VLA models remains a fundamental and open challenge.

To this end, we propose **Long-VLA**, the first end-to-end VLA model specifically designed for long-horizon robotic manipulation. Our approach introduces an input-level adaptation strategy that segments each subtask into a moving phase and an interaction phase, applying phase-aware input masking to guide learning. Phase-aware masking leverages third-person views in moving phases and egocentric views for precise manipulation, ensuring the model focuses on phase-relevant cues, reduces representation shifts, and improving skill chaining. Importantly, this strategy preserves end-to-end training and lets the policy network exploit large-scale data with minimal input changes. Additionally, this method introduces a scalable, architecture-agnostic module that integrates seamlessly into existing VLAs without altering their core structure. Finally, we present L-CALVIN and show that Long-VLA outperforms state-of-the-art methods on simulated and real-world robotic tasks, with robust performance on diverse long-horizon tasks.

## 2 Related Work

**Vision-Language-Action Models.** Vision-Language-Action (VLA) models [25, 26, 27, 28, 29, 30, 31, 32, 33, 34, 35, 36] integrate visual perception, language understanding, and robotic action generation to enable autonomous control from multimodal inputs. They represent a promising paradigm for learning generalist policies, largely driven by pretraining on increasingly large and diverse robot learning datasets [37, 38, 39, 40]. However, existing VLA models are designed for short-horizon tasks, where the semantic and action spaces are relatively limited and well-structured. This leaves a gap in their ability to generalize to long-horizon scenarios.

**Long-horizon Robot Manipulation.** Long-horizon manipulation is typically approached via decomposition [41, 42, 43, 44, 45, 46, 47, 48, 49], where complex tasks are decomposed into subtasks with separate local policies optimized for each subgoal. However, such modular approaches lack explicit modeling of transitions and dependencies between subtasks, leading to the skill-chaining challenge [18, 50], where dynamic coupling and error propagation across stages can degrade overall performance. See Appendix A.2 for details. To address this, recent work has focused on two primary directions [20, 21, 22, 19, 3, 23, 24]: (1) Online adaptive optimization, which addresses error propagation during execution through online fine-tuning and trial-and-error strategies, such as real-time correction [21] or reward adjustment [22, 51]; and (2) Minimizing the train-test gap, which aims to reduce discrepancies between training and deployment to improve robustness. A representative example is Plan-Seq-Learn [3], which decouples motion planning and execution using different input modalities to mitigate skill-chaining effects.

**Long-horizon Robot Manipulation in VLA Models.** Recent approaches such as Dex-VLA [17] and $\pi_0$ [16] introduce task decomposition into the VLA framework, leveraging LLMs to simplify subtask execution by reducing learning complexity. However, current VLA methods have yet to consider skill chaining from long-horizon manipulation. A key challenge is that reward-based online methods are incompatible with the offline training paradigm of VLA models, where reward signals are typically unavailable. Modular architectures split tasks into separate modules, hindering joint training and contradicting VLA's end-to-end paradigm. To bridge this gap, we aim to address the challenges of skill chaining for long-horizon tasks while maintaining the scalability and data. Then we propose Long-VLA, the first end-to-end VLA model designed for long-horizon manipulation.

## 3 Method

### 3.1 Revisiting Decomposition Strategy

Before introducing our method, we first investigate whether decomposition is essential for VLA models. Our intuition is to further divide each subtask into two fine-grained phases: a moving phase and an interaction phase. Prior work [3] used inverse kinematics (IK) for precise motion trajectories, but accurate 3D targets and practical IK are often infeasible. We instead train a dedicated moving policy to replace IK.

**Decomposed Data Collection.** To assess the feasibility of phase-level decomposition, a preliminary study is conducted on the CALVIN dataset [40]. From the original trajectories, a new dataset termed L-CALVIN is constructed by segmenting each task into movement and interaction phases. The interaction phase is handled by a pre-trained VLA model, while a separate moving policy is trained on on movement-phase data. We extract 64-frame sequences labeled via the task detector in [40]. Language instructions are augmented with movement-specific commands based on detected objects and locations. To ensure phase alignment, the cutting point is set 10–15 frames prior to the object's state change.

Table 1: Comparison between MDT and MDT enhanced with a Moving Policy (MP) across different task horizons.

| Method | Tasks Completed in Sequence (%) | | | | |
| --- | --- | --- | --- | --- | --- |
| | 1 | 2 | 3 | 4 | 5 |
| MDT [52] | 93.3 | 82.4 | 71.9 | 60.9 | 51.1 |
| MDT + MP | 95.8 | 91.7 | 87.5 | 66.7 | 54.2 |

**Performance with Decomposition Strategy.** As shown in Table 1, combining MDT [52] with a separate moving policy leads to a clear performance improvement, demonstrating the effectiveness of our decomposition strategy. However, training two separate models is suboptimal for scalable long-horizon learning. To address this limitation, we propose Long-VLA, a unified end-to-end VLA model that leverages phase-specific data more effectively.

### 3.2 Long-VLA

#### 3.2.1 Training Paradigm

**Data and Phase Decomposition.** As described in Section 3.1, each language-annotated trajectory is decomposed into $\tau = \left[(s_t^{\mathcal{M}}, a_t^{\mathcal{M}}) t \in [0, d], (s_t^{\mathcal{I}}, a_t^{\mathcal{I}}) t \in [d+1, T]\right]$, where $\mathcal{M}$ and $\mathcal{I}$ denote the moving and interaction phases, respectively, and $d$ is the cutting point time step. To enable training and inference within a unified end-to-end VLA framework, we extend the original action representation by adding a one-dimensional phase identifier $s_p$, which indicates the current phase. The final action token is represented as $[x, y, z, eu_x, eu_y, eu_z, s_g, s_p]$, where $(x, y, z)$ are the Cartesian coordinates of the robot end-effector, $(eu_x, eu_y, eu_z)$ denote its Euler-angle-based orientation, and $s_g$ indicates the gripper state (open or closed). The phase identifier $s_p$ is set to $-1$ during the moving phase and $1$ during the interaction phase. During inference, $s_p$ is initialized to $-1$.

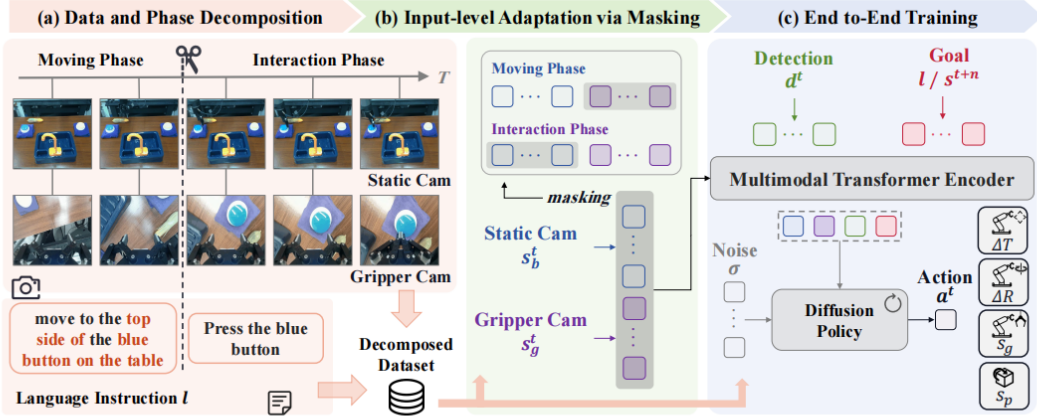

Figure 2: Overview of Long-VLA. (a) Task decomposition with aligned visual observations and language annotations. (b) Phase-aware masking enables the model to selectively attend to relevant tokens during attention computation without modifying the input structure. (c) End-to-end training using decomposed data with phase-aware masking.

**Input-level Adaptation Strategy via Masking.** We argue that during the moving phase, the model should focus on precise object navigation using third-person camera views, as the gripper camera view is minimally informative at this stage. In contrast, during the interaction phase, attention should shift to the gripper camera to mitigate visual distribution shifts and enable accurate manipulation. Based on these observations, we propose an input-level adaptation strategy that dynamically adjusts visual inputs according to the current task phase. To dynamically adjust different visual inputs across task phases, we adopt a masking strategy rather than directly removing entire modalities. Specifically, each token is assigned a binary mask $\mathbf{m} \in \{0,1\}^N$, where $\mathbf{m}_i = 1$ indicates that the $i$-th token participates in the attention computation, and $\mathbf{m}_i = 0$ otherwise. The binary vector $\mathbf{m}$ is then expanded into an attention mask matrix $\mathbf{M} \in \{0,1\}^{N \times N}$, with each element defined as: $\mathbf{M}_{ij} = \mathbf{m}_i \cdot \mathbf{m}_j$. This ensures that attention is only computed between active token pairs. Given the query-key similarity matrix $\mathbf{P} \in \mathbb{R}^{N \times N}$, computed as $\mathbf{P} = \mathbf{Q}\mathbf{K}^T/\sqrt{C}$, the masked attention weights $\mathbf{A}$ are calculated as:

$$\mathbf{A}_{ij} = \frac{\exp(\mathbf{P}_{ij})\mathbf{M}_{ij}}{\sum_{k=1}^{N}\exp(\mathbf{P}_{ik})\mathbf{M}_{ik}}, \quad \text{for } 1 \leq i, j \leq N. \tag{1}$$

By applying this masking strategy, the model selectively focuses on relevant tokens during attention computation without altering the input structure, thereby preserving modality consistency while adapting to different task phases. Details of the masking strategy are provided in Appendix ??.

**Training Loss.** For action generation, we employ a conditional diffusion model to generate $a_t$. Using the decomposition dataset, the model is trained with a single score matching loss that jointly supervises both the moving and interaction phases:

$$\mathcal{L}_{\text{Diff}} = \mathbb{E}_{a \sim p_{\text{data}}} \mathbb{E}_{n \sim \mathcal{N}(0, \sigma^2 I)} \| D_\theta(\tilde{a}_t, e_{post}, \sigma_t) - a_t \|_2^2, \tag{2}$$

where action denoiser $D_\theta(\tilde{a}_t, e_{post}, \sigma_t)$ that progressively refines noisy actions $\tilde{a}_t$, $\sigma_t$ is noise level using multi-modal tokens. To ensure that the visual goals are semantically consistent with language instructions, we employ an InfoNCE loss $\mathcal{L}_{\text{Goal}}$, as detailed in Appendix C.2. As a result, the total training loss is formulated as:

$$\mathcal{L} = \mathcal{L}_{\text{Diff}} + \alpha \mathcal{L}_{\text{Goal}}. \tag{3}$$

where $\alpha$ is a hyperparameter, which we set to 0.1.

#### 3.2.2 Model Achitecture

Long-VLA policy $\pi_\theta(a^t \mid s^t, d^t, g)$ predicts the action $a^t$ conditioned on the current observation $s^t$, the detection input $d^t$ associated with $s^t$, and the latent goal $g$, where $t$ denotes the timestep.

**Observation Encoder.** The observation $s^t$ includes the gripper camera view $s_g^t$ and the static camera view $s_b^t$, which are embedded into $e_g$ and $e_b$, respectively, using a trainable ResNet-18 encoder [53].

**Goal Encoder.** To leverage the unlabeled play data, we follow a strategy similar to [52], where the future observation $s^{t+n}$ is used as a visual goal in the absence of language instructions, and a language annotation is used as the goal when available. Both types of goals are encoded using the text and image encoders of the frozen CLIP model [54], resulting in $e_{goal}^l$ and $e_{goal}^o$, respectively.

**Detection Integration.** To support accurate object navigation and interaction in dynamic scenes, we incorporate additional detection information. Specifically, we fine-tune Grounding DINO [55] with LoRA [56] on a subset of the CALVIN dataset to achieve reliable, fine-grained object localization. The model $f_d$ predicts pixel-level bounding boxes from third-person images conditioned on language queries. These bounding boxes are projected into the latent space using a trainable positional encoder to obtain detection features $e_d$. We then modulate the static camera features using $e_d$ via a FiLM strategy [57], resulting in a detection-enhanced representation $\hat{e}_b$. Further implementation details are provided in Appendix C.1.

**Multimodal Encoder.** The multi-modal encoder in our model is based on a GPT-2-style Transformer architecture. The input $e_{pre}$ is defined as $[\hat{e}_b; e_g; e_{goal}; e_d]$, which concatenates all modality features and encodes them into latent perceptual tokens $e_{post}$.

**Action Decoder.** We employ a conditional diffusion model to generate actions $a_t$ by progressively denoising from Gaussian noise, with the reverse process implemented using DDIM sampling: $x_{t-1} = \frac{\sigma_{t-1}}{\sigma_t} x_t - (-h)e^{-1}\hat{x}_0$, where $\hat{x}_0$ is the denoised prediction. After decoding through the diffusion model, the output is mapped to action vectors via a two-layer MLP with GELU activation.

## 4 Experiment

In this section, we address the following research questions: **RQ1**: How does our proposed paradigm enhance the base policy? **RQ2**: How does our Long-VLA compare with state-of-the-art (SOTA) methods? **RQ3**: What are the key design components of our Long-VLA?

### 4.1 Experiments Setup

**Simulation & Real-world Experiment.** We select CALVIN as our simulation platform due to its focus on long-horizon tasks, and introduce L-CALVIN, a new benchmark that extends task sequences from 5 to 10 steps based on CALVIN's data protocols (see Appendix B.1). In addition, we design two real-world tasks: (1) sequentially placing blocks into a bowl (sequence length 8), and (2) a complex kitchen cleaning task (sequence length 4). The first task emphasizes longer temporal dependencies, while the second evaluates complex action execution. This setup enables a comprehensive assessment of long-horizon performance.

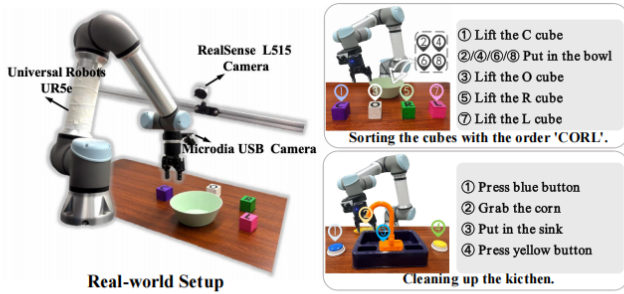

Figure 3: Real-world setup.

**Base Policy & Other Baselines.** In simulation and real-world environments, we select MDT [52] as our base policy. This choice is primarily motivated by MDT's strong ability to process multimodal inputs, as demonstrated by its performance in the CALVIN environments. In real-world settings, this decision is further supported by MDT's data efficiency, particularly its ability to leverage unlabeled in-domain data to improve model performance even without text annotations. In addition, we include several baselines to more comprehensively evaluate the effectiveness of our method: video generation-based VLA models (GR-1 [58] and UP-VLA [59]) and VLM-based VLA models (RoboVLMs [60], VLAS [31], and OpenVLA [1]). Since $\pi_0$ [16] is not evaluated in the CALVIN environment, we use it as a baseline in real-world experiments.

| | | Rotate blue block right | Close drawer | Push red block right | Push red block left | Rotate red block right | Move slider right | Push blue block left | Rotate red block left | Lift pink block slider | |
|---|---|---|---|---|---|---|---|---|---|---|---|

| Train→Test | Method | \multicolumn{10}{c}{Tasks Completed in Sequence} |
|---|---|---|---|---|---|---|---|---|---|---|---|
| | | 1 | 2 | 3 | 4 | 5 | 6 | 7 | 8 | 9 | 10 |
| D→D | Base Policy | 0.86 | 0.64 | 0.53 | 0.47 | 0.37 | 0.31 | 0.28 | 0.21 | 0.13 | 0.11 |
| | Long-VLA | **0.92** (7% ↑) | **0.74** (16% ↑) | **0.65** (23% ↑) | **0.50** (28% ↑) | **0.43** (16% ↑) | **0.39** (26% ↑) | **0.36** (29% ↑) | **0.30** (42% ↑) | **0.26** (100% ↑) | **0.20** (81% ↑) |
| ABCD→D | Base Policy | 1.00 | 0.95 | 0.93 | 0.86 | 0.82 | 0.75 | 0.68 | 0.61 | 0.53 | 0.45 |
| | Long-VLA | **1.00** (0% ↑) | **1.00** (5% ↑) | **0.98** (5% ↑) | **0.91** (6% ↑) | **0.85** (4% ↑) | **0.82** (10% ↑) | **0.79** (16% ↑) | **0.70** (15% ↑) | **0.63** (19% ↑) | **0.56** (25% ↑) |

Figure 4: Simulation performance on L-CALVIN.

| | | Lift the C cube | Put in the bowl | Lift the O cube | Put in the bowl | Lift the R cube | Put in the bowl | Lift the L cube | Put in the bowl |
|---|---|---|---|---|---|---|---|---|---|

| Unseen Type | Method | \multicolumn{8}{c}{Tasks Completed in Sequence} |
|---|---|---|---|---|---|---|---|---|---|
| | | 1 | 2 | 3 | 4 | 5 | 6 | 7 | 8 |
| Random Localization | Base Policy | 0.70 | 0.60 | 0.40 | 0.35 | 0.20 | 0.05 | 0 | 0 |
| | Long-VLA | **0.95** (35% ↑) | **0.95** (58% ↑) | **0.85** (112% ↑) | **0.80** (128% ↑) | **0.50** (150% ↑) | **0.50** (900% ↑) | **0.50** (+∞ ↑) | **0.45** (+∞ ↑) |
| Unseen Lighting | Base Policy | 0.50 | 0.40 | 0.35 | 0.30 | 0.15 | 0.05 | 0.00 | 0.00 |
| | Long-VLA | **0.80** (60% ↑) | **0.75** (87% ↑) | **0.65** (85% ↑) | **0.55** (83% ↑) | **0.45** (200% ↑) | **0.30** (700% ↑) | **0.30** (+∞ ↑) | **0.25** (+∞ ↑) |
| Visual Distraction | Base Policy | 0.55 | 0.45 | 0.25 | 0.20 | 0.05 | 0.00 | 0.00 | 0.00 |
| | Long-VLA | **0.85** (54% ↑) | **0.80** (77% ↑) | **0.70** (180% ↑) | **0.65** (225% ↑) | **0.50** (900% ↑) | **0.45** (+∞ ↑) | **0.40** (+∞ ↑) | **0.35** (+∞ ↑) |

Figure 5: Real-world Performance on Sorting.

### 4.2 Long-VLA v.s. Base Policy

**Evaluation on Simulation Scene.** As shown in Figure 4, our model achieves performance improvements in the D→D and ABCD→D of the L-CALVIN benchmark. We observe that as the task duration increases, the performance improvement for tasks becomes more significant. Moreover, in scenarios with limited data, our approach yields even greater improvements, which is particularly meaningful for applications constrained by data availability. It should be noted that the above observations are based on the seen scenario. To further evaluate the performance in unseen scenarios, we additionally conducted tests in real-world environments with previously unseen settings.

**Evaluation on Real-World Scene (Sorting).** To evaluate the model's basic performance on long-horizon tasks, we employ the sorting task as the testing scenario, which is relatively simple, consisting of only a few cubes and a bowl, which results in minimal visual redundancy. Under such conditions, the sorting task serves as an effective means to assess the model's fundamental capabilities in long-horizon task execution. As shown in Figure 5, while the success rate of the base policy drops to zero after the seventh task, our approach is still able to achieve a success rate of nearly 25% across all eight tasks. Furthermore, when examining the improvement ratio for each task, we observe that the relative performance gain of our model increases with task sequence length. This demonstrates the robustness of our method in handling long-horizon tasks.

**Evaluation on Real-World Scene (Cleaning).** To provide a more rigorous evaluation, we propose a cleaning task featuring a wider variety of actions—such as pressing, grabbing, and placing—and increased visual complexity with more distracting information. This task serves as a more challenging and representative benchmark for assessing long-horizon performance. As shown in Figure 6, our model achieves significant improvements over the base policy across all time horizons. Notably, in the cleaning scenario, these improvements are even more pronounced than the corresponding values in the sorting scenario. These results demonstrate that our method is particularly effective in handling visual distractions and complex environments, which we attribute to our proposed long-horizon adaptation paradigm.

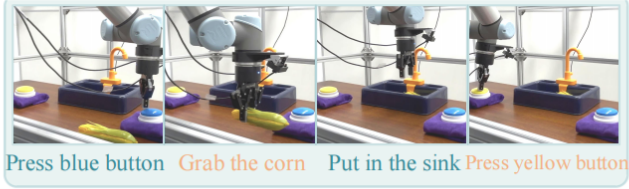

| Unseen Type | Method | Tasks Completed in Sequence | | | |
|---|---|---|---|---|---|
| | | 1 | 2 | 3 | 4 |
| Random Localization | Base Policy | 12/20 | 8/20 | 5/20 | 3/20 |
| | Long-VLA | **18/20** (50% ↑) | **14/20** (75% ↑) | **13/20** (160% ↑) | **11/20** (226% ↑) |
| Unseen Lighting | Base Policy | 9/20 | 6/20 | 5/20 | 2/20 |
| | Long-VLA | **16/20** (77% ↑) | **13/20** (116% ↑) | **12/20** (140% ↑) | **9/20** (250% ↑) |
| Visual Distraction | Base Policy | 11/20 | 7/20 | 6/20 | 3/20 |
| | Long-VLA | **17/20** (54% ↑) | **13/20** (85% ↑) | **12/20** (100% ↑) | **11/20** (266% ↑) |

Figure 6: Real-world performance on cleaning.

### 4.3 Long-VLA v.s. SOTA

As presented in Table 2 and Figure 7, our model achieves the best performance in both simulated and real-world experiments. In real-world robotic experiments, our method consistently outperforms the state-of-the-art algorithm $\pi_0$ across the generalization task.

Table 2: Comparison with SOTA methods on L-CALVIN simulation benchmark.

| Train→Test | Method | Tasks Completed in Sequence | | | | | | | | | | Avg. Len |
|---|---|---|---|---|---|---|---|---|---|---|---|---|
| | | 1 | 2 | 3 | 4 | 5 | 6 | 7 | 8 | 9 | 10 | |
| | GR-1 | 0.83 | 0.58 | 0.48 | 0.35 | 0.24 | 0.17 | 0.13 | 0.09 | 0.05 | 0.04 | 2.96 |
| D→D | RoboVLMs | 0.81 | 0.60 | 0.44 | 0.34 | 0.28 | 0.15 | 0.10 | 0.08 | 0.05 | 0.03 | 2.88 |
| | **Long-VLA** | **0.92** | **0.74** | **0.65** | **0.50** | **0.43** | **0.39** | **0.36** | **0.30** | **0.26** | **0.20** | **4.75** |

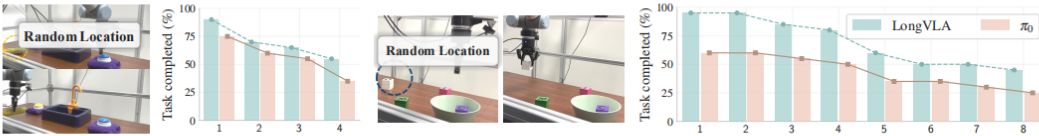

Figure 7: Comparison with SOTA method on real-world scenarios. (Left: cleaning; Right: sorting)

These performance gains stem from two key factors: the robust capability of our base policy and the substantial enhancement provided by our proposed long-term strategy. The experimental results clearly validate that our solution effectively elevates overall performance beyond current SOTA approaches. These findings demonstrate that while foundational capabilities may suffice for single-step tasks, long-horizon tasks demand minimal error accumulation. Our model addresses these challenges more effectively than existing approaches. More experiments can be found in Appendix D.6.

### 4.4 Ablation Analyses

We validate the key design elements of Long-VLA —decomposition strategy, input-level adaptation, and unified model-in Table 3. For comparison and evaluation, the **Base Policy** uses only a unified model, while **Long-VLA** incorporates all three strategies to achieve better performance.

Table 3: Ablation of key components. "Dec.", "Inp.", and "Uni." represent Decomposition Strategy, Input-level Adaptation, and Unified Model, respectively. The values in the table represent the Avg. Len under different test scenarios.

| Dec. | Inp. | Uni. | Real (Sorting) | Real (Cleaning) | Sim (D-D) |
|---|---|---|---|---|---|
| ✗ | ✗ | ✓ | 2.3 | 1.4 | 4.11 |
| ✓ | ✗ | ✓ | 3.6 (1.3 ↑) | 1.7 (0.3 ↑) | 4.42 (0.31 ↑) |
| ✓ | ✓ | ✗ | 4.1 (1.8 ↑) | 2.0 (0.6 ↑) | 4.76 (0.65 ↑) |
| ✓ | ✓ | ✓ | **5.5** (3.2 ↑) | **2.8** (1.4 ↑) | **4.81** (0.70 ↑) |

**Ablation on Decomposition Strategy.** As shown in Table 3, introducing the decomposition strategy to the Base Policy leads to a significant improvement. This enhancement can be mainly attributed to the decomposition mechanism, which effectively mitigates the negative impact of imperfect prior executions by enabling more precise movements. This observation is fully consistent with the findings in RH20T-P [61], suggesting that adopting a decomposition strategy allows the model to efficiently correct task execution based on updated scene information when deviations occur.

**Ablation on Input-level Adaptation.** Performance significantly improves with input-level adaptation, mainly from adding detection data during movement for better control and removing unwanted third-person visual interference during interaction for enhanced robustness. While this improves generalization somewhat, it is important to note that the different input modalities between stages prevent unified training, reducing available data per stage and limiting overall performance gains.

**Ablation on Unified Model.** Finally, introducing our masking mechanism achieves the best performance. This is because our approach not only retains the visual generalization capability provided by input-level adaptation but also enables joint training within the VLA model. In this manner, a shared policy can be trained across stages, effectively combining the data-driven advantages of end-to-end VLA models with strategies for long-horizon tasks, significantly enhancing performance.

### 4.5 Scalability of Long-VLA Paradigm

We demonstrate the versatility of the Long-VLA paradigm by evaluating it with multiple base architectures. While our main experiments use MDT [52] as the backbone, we also report results with HULC [62]. As shown in Table 4, Long-VLA consistently achieves strong performance across different models, highlighting its effectiveness and architecture-agnostic design. This versatility not only validates the robustness of our approach but also highlights its potential for integration with a wide range of existing VLA models, paving the way for more capable and adaptable robotic systems in the future.

Table 4: The Versatility of Long-VLA in Sim(D→D).

| Method | Avg. Len | Method | Avg. Len |
|---|---|---|---|
| Base Policy (HULC) | 2.65 | Base Policy (MDT) | 4.11 |
| Long-VLA (HULC) | 3.30 (0.65 ↑) | Long-VLA (MDT) | 4.81 (0.7 ↑) |

## 5 Conclusion

In conclusion, Long-VLA advances the field of generalist robotics by introducing a unified, end-to-end approach to long-horizon manipulation, effectively addressing the persistent challenge of skill chaining through a novel phase-aware input adaptation strategy. By segmenting each subtask into movement and interaction phases with targeted masking, Long-VLA mitigates distribution shifts and enhances subtask compatibility, enabling robust performance across complex tasks. Our extensive evaluations on the L-CALVIN benchmark and real-world scenarios demonstrate that Long-VLA not only surpasses existing state-of-the-art methods but also offers a scalable, architecture-agnostic solution that can be easily integrated into current VLA models, paving the way for more capable and adaptable robotic systems in the future.

## Limitation

While our experiments confirm the effectiveness of the decomposition method in enhancing the performance of VLAs on long-horizon tasks, our approach has several limitations that could be addressed in future work. Firstly, the decomposition of training datasets for different phases still depends on manual effort, which could potentially be automated with the assistance of VLMs. Additionally, the scope of long-horizon tasks we have considered is still limited. Although our model can reduce the initial state gap, it requires further refinement to handle failure cases in longer sequences. Moreover, the scope of our tested sequence lengths is constrained. While our model mitigates the initial state gap, it does not address execution failures under precise initial conditions. Furthermore, we aim to diversify the migratory compatibility of model frameworks in future investigations. We hope that our research will pave the way for more practical and reliable methods, providing deeper insights into long-horizon tasks.

## Acknowledgments

This work was supported by the National Science and Technology Innovation 2030 - Major Project (Grant No. 2022ZD0208800), NSFC General Program (Grant No. 62176215), National Natural Science Foundation of China (Grant No. U21A20485), and the Fundamental Research Funds for Xi'an Jiaotong University (Grant No. xzy022024012).

## A  Preliminaries

### A.1  Definition of VLA Models

Imitation learning with language instructions enables an agent to learn manipulation actions from a dataset of $N$ trajectories $\{\tau_i\}_{i=0}^{N}$. Each trajectory $\tau$ consists of a sequence of state-action pairs, $\tau = \{(s_t, a_t)\}_{t=0}^{T}$, where $s_t$ and $a_t$ denote the state and action at time step $t$, respectively. The learning objective follows the standard imitation learning paradigm [?], which maximizes the log-likelihood of the demonstrated actions conditioned on states and language instructions:

$$\mathcal{L} = \mathbb{E}_{(\tau,l)\sim\mathcal{D}} \left( \sum_{t=0}^{|\tau|} \log \pi_\theta(a_t|s_t, l) \right), \quad (4)$$

where $\mathcal{D} = \{(\tau, l)_i\}_{i=0}^{N}$ denotes the dataset of trajectory-instruction pairs. As illustrated in Figure 8, the policy $\pi_\theta$ is modeled as a text-conditioned action predictor, mapping from the current state and language instruction to the action space:

$$\pi_\theta(a|s, l) : S \times L \to A, \quad (5)$$

where $S$, $L$, and $A$ denote the state, language instruction, and action spaces, respectively. Through optimizing $\pi_\theta$, the agent learns to execute actions that align with both the observed states and the given language commands.

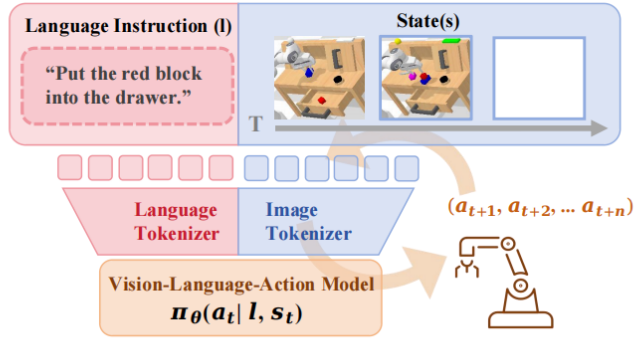

Figure 8: Definition of VLA Models. VLA models generate sequences of actions conditioned on input language instructions and the current environmental state.

### A.2 Skill Chain Challenges in Long-horizon Tasks

As is shown in Figure 9, we verify the skill chain challenage also exits in current VLA models. In subfigure (b), we can see that even after removing the interference from previous tasks, the success rate for individual tasks still continues to decline. In subfigure (c), we further verify this point. We conducted a single-task test, allowing the model to be tested independently in both the Independent and Continuous environments. We found that once the tasks are performed continuously, the model's single-task performance is lost. The above experimental analyses all demonstrate this issue.

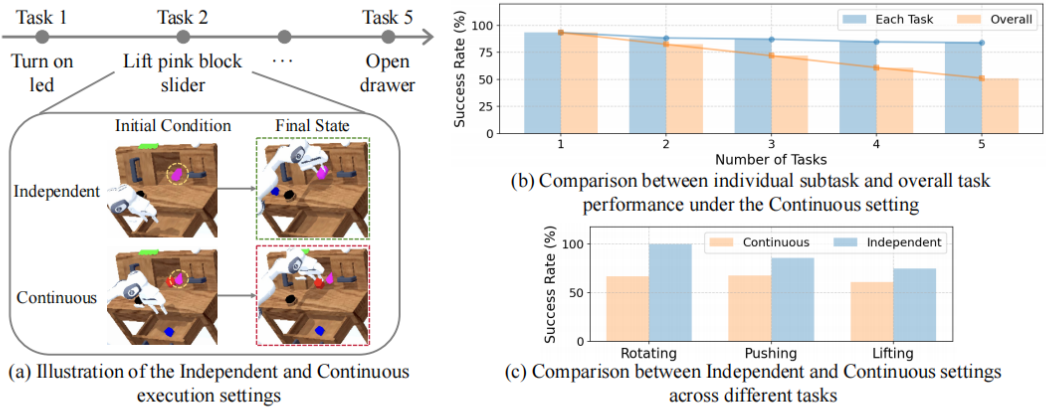

Figure 9: (a) Illustration of skill-chaining challenges like state mismatch in CALVIN benchmark. In the independent setting, each subtask starts from a state within the training distribution. In the continuous setting, subtasks are executed sequentially, causing potential distribution shifts. (b) The performance drop in individual subtasks suggests the potential impact of state mismatch. (c) Validates the effect of state mismatch (e.g., position differences) across different tasks.

## B  L-CALVIN

### B.1  L-CALVIN.

**Motivation of L-CALVIN.** Improving the performance of long-horizon tasks is crucial for robotic systems operating in real-world environments, where robots must execute continuous and complex operations. In such settings, the quality of task execution at each stage directly influences subsequent tasks, making long-horizon task planning and execution inherently more challenging than single-task assessments.

Table 5: Language instruction examples for the movement and interaction phases. We extend the movement phase instructions to increase diversity, while retaining the original CALVIN instructions for the interaction phase.

| Task Type | `rotate red block right` | `move slider left` | `open drawer` | `turn on led` |
|---|---|---|---|---|
| Movement Instruction | move: move to the red block from the top | move: go to the slider from the right | move: move in front of the closed drawer | move: move to the top side of the led button |
| | move: go to the top side of the red block | move: reach the the right side of slider handle | move: go to the drawer handle | move: approach the led button from the top |
| Interaction Instruction | Take the red block and rotate it to the right | Push the sliding door to the left side | Pull the handle to open the drawer | press the button to turn on the led light |

**Algorithm 1** CALVIN Sequence Generation in original paper [40]
**Input**: $A$: action sequence
$Aset$: action sequence set
$S$: start state
$len$: sequence length
**Initialize**: CheckSequence($S$, $A$)
1: Random $A$ with $len$ tasks
2: **if** CheckSequence($S$, $A$) **then**
3:     $Aset$.append($A$)
4: **end if**

**Algorithm 2** L-CALVIN Sequence Generation
**Input**: $A$: action sequence
$Aset$: action sequence set
$S$: current state
$len$: sequence length
**Initialize**: RandomPossibleNext($S$)
1: Random $S$
2: **for** $i$ in range($len$) **do**
3:     **if** PossibleNext($S$) **then**
4:         $N \leftarrow$ RandomPossibleNext($S$)
5:         $A$.append($N$)
6:     **else**
7:         break
8:     **end if**
9: **end for**
10: $Aset$.append($A$)

**Background of CALVIN.** The CALVIN benchmark provides a long-horizon evaluation environment composed of sequences of five decomposed subtasks. It includes four different scene splits (A, B, C, D), with evaluations typically conducted in scene D. In CALVIN, 34 individual tasks are grouped into 11 major categories to generate random evaluation sequences, which limits the maximum sequence length to 11 tasks. Additionally, sequence generation follows a random sampling and verification process, which is both inefficient and unable to scale to longer horizons.

**Limitation of CALVIN.** Existing evaluation protocols primarily focus on short temporal horizons, typically involving sequences of only five tasks. This limited scope fails to adequately capture the performance of robotic systems under more realistic, extended operation scenarios. When faced with longer task sequences, current models experience significant performance degradation, with different models' performances eventually converging to similar levels, indicating a model-agnostic challenge.

**Overview of L-CALVIN Benchmark.** To address this limitation, we introduce L-CALVIN, an extended version of the CALVIN benchmark. L-CALVIN increases the task sequence length from five to ten consecutive tasks, providing a more comprehensive benchmark for evaluating long-horizon task performance. This extended evaluation framework allows for a more rigorous examination of task execution consistency, adaptability, and long-term efficiency. L-CALVIN removes the category-based constraint by treating all 34 tasks as independent categories, enabling the generation of sequences longer than ten tasks. Furthermore, each task execution is decomposed into two distinct phases: *movement* and *interaction*, as illustrated in Figure 15. By switching camera perspectives between phases, we enhance the agent's semantic understanding and mitigate the state initialization gap between tasks.

**Sequence Generation.** Corresponding to the L-CALVIN benchmark, we construct a phase-specific dataset to support policy learning under diverse language instructions. Using ground truth information from the CALVIN simulation, we re-annotate the dataset by decomposing trajectories based on object movement states. We introduce additional language instructions specifically for the movement phase, while keeping interaction phase instructions consistent with the original annotations. During validation, we use language expressions that differ from those seen during training to assess generalization. Examples of the augmented language instructions are presented in Table 5. For phase decomposition, we detect position changes of target objects between consecutive frames to segment movement and interaction phases. Compared to the random sequence generation approach of CALVIN (Algorithm 1), L-CALVIN adopts a structured generation method (Algorithm 2) that incrementally selects executable tasks, allowing efficient construction of longer sequences without post-validation.

**Overview of L-CALVIN Dataset.** We randomly extract sequences with a window size of 64 frames. Each sequence is labeled if a task detector identifies task completion, following the approach in [40]. The associated language instruction is then augmented with a movement command using a manually designed template based on the identified object and location, while retaining the original instruction as the interaction phase. Based on the task prompt, we determine the cutting point to be 10 to 15 frames before the object's state change to ensure coherent visual composition. In total, we retain language annotations for 34 tasks, resulting in 372 unique instructions.

**Comparison with CALVIN.** The original CALVIN approach ensures that no two tasks from the same major category appear within a sequence; however, this constraint limits sequence length and reduces data generation efficiency. In contrast, L-CALVIN leverages a finer-grained categorization at the minor task level, significantly increasing the diversity of feasible sequences and approximately tripling the number of block manipulation tasks. This enhanced structure makes L-CALVIN particularly suited for evaluating long-horizon task execution under realistic and challenging conditions.

## C Model Details

### C.1 Detection Module Details

To enable accurate object navigation and interaction in dynamic scenes, we incorporate additional detection information. Specifically, we select a subset of images from the CALVIN dataset and fine-tune Grounding DINO [55] (Swin-OGC variant) using LoRA [56]. The model, denoted as $F_b$, is trained to localize target objects based on language queries $l$. Given a third-person image $o_s$ from the movement phase, $F_b$ outputs a set of bounding boxes:

$$o_b = F_b(l, o_s), \quad (6)$$

where $o_b \in \mathbb{R}^{N_b \times 4}$. This effectively transforms image-based object representations into fine-grained, pixel-level spatial information. To align these spatial locations with the model's latent representation, we apply a trainable positional encoding function $\phi$ to project the bounding boxes into the feature space:

$$e_b = \phi(o_b), \quad e_b \in \mathbb{R}^{2N_b \times d_v}, \quad (7)$$

where $e_b$ represents the latent feature encoding of the detection boxes. We then integrate this detection-aware feature into the visual embedding $e_s$ of the static camera view using the FiLM mechanism [57]. This is achieved through two learnable projection matrices, $W_{\text{mult}}$ and $W_{\text{add}}$, initialized to zero for unbiased fusion:

$$\hat{e}s = (1 + W_{\text{mult}}(e_b)) \odot e_s + W_{\text{add}}(e_b), \quad (8)$$

where $\odot$ denotes element-wise multiplication. This formulation allows dynamic modulation of image features based on detection results. Finally, we construct the multimodal input embedding $e_p re$ for the transformer encoder by concatenating the goal feature $e_{goal}$, the gripper camera observation encoding $z_g$, the detection encoding $e_d$, and the FiLM-enhanced static view $\hat{e}_s$:

$$e_{pre} = [\hat{e}_s;, e_g;, e_{goal};, e_d]. \quad (9)$$

## C.2 Alignment Loss

To align the image goal and language goal representations, we utilize a contrastive learning loss, which can be formulated as:

$$\mathcal{L}_\text{s} = (1 - \frac{1}{2B}) \sum_{t=1}^{B} (\log \left( \frac{\exp\left(S(z_t^o, z_t^l)/\nu\right)}{\sum_{n=1}^{B} \exp\left(S(z_t^o, z_n^l)/\nu\right)} \right) \tag{10}$$

where $S(\cdot, \cdot)$ denotes the cosine similarity, $\nu$ is a temperature parameter, and $B$ is batch size.

## D Experiment Details

### D.1 Baseline.

We conduct supplementary experiments with additional baseline models to facilitate a more comprehensive comparison of different types of models.

- **GR-1 [58]:** A GPT-style unified transformer that leverages large-scale language-conditioned video generative pre-training and is then fine-tuned on robot data to jointly predict future frames and continuous actions, boosting CALVIN success to 94.9% and exhibiting strong zero-shot generalization to unseen scenes, objects, and real-world tasks.
- **RoboVLMs [60]:** A flexible framework that turns off-the-shelf Vision–Language Models into VLA policies, systematically dissecting backbone selection, policy formulation (history-aware continuous actions with a policy head), and timing for cross-embodiment data.
- **VLAS [31]:** An end-to-end VLA policy that fuses raw speech, images, and actions without an external automatic speech recognition, aligns speech inside LLaVA, and adds a voice–retrieval RAG for user-specific knowledge.
- **$\pi_0$ [16]:** A cross-embodiment VLA model that augments a PaliGemma backbone with a flow-matching action expert to generate high-frequency continuous action chunks; pre-trained on 10,000 h of diverse single- and dual-arm data and post-trained on curated demos.

### D.2 Implementation Details.

**Hardware of Real-world Experiments.** As shown in Figure 3, we utilizes a UR5e robotic arm within a real-world desktop environment, which includes bowls, cubes, corn, and a sink. We employ two cameras: one positioned diagonally in front of the robotic arm to provide an overhead view, and another mounted on the robot gripper for precise interactions with the objects.

**Real-world Play Dataset.** We collected 200 demonstrations for each of the Sorting and Cleaning tasks via teleoperation. By pressing the 'm' key, we performed data decomposition during collection. In the Sorting task, blocks were placed arbitrarily across the tabletop. For the Cleaning task, while constraining the buttons and sink so that they could only move along the front–back axis, the position of the corn was randomized on the tabletop. In addition, we gathered approximately two hours of unlabeled play data in which volunteers freely explored the tabletop arranged with all experimental objects.

**Detection Information.** Since the CALVIN environment differs significantly from the object states in real-world scenarios, and the captured images have relatively low resolution, we fine-tune GroundingDINO within to obtain detection information. However, in real-world experiments, we zero-shot use GroundingDINO on images, with prompts related to the task target objects, such as "the corn" or "the blue button".

**Training Details.** For the L-CALVIN task, we train the D→D model for 40 epochs using a learning rate of 1e-4 and a batch size of 128, taking approximately 18 hours on four NVIDIA A100 GPUs (80GB VRAM). For the real-world tasks, we train one model for each of the two tasks over 800 epochs, taking about 28 hours.

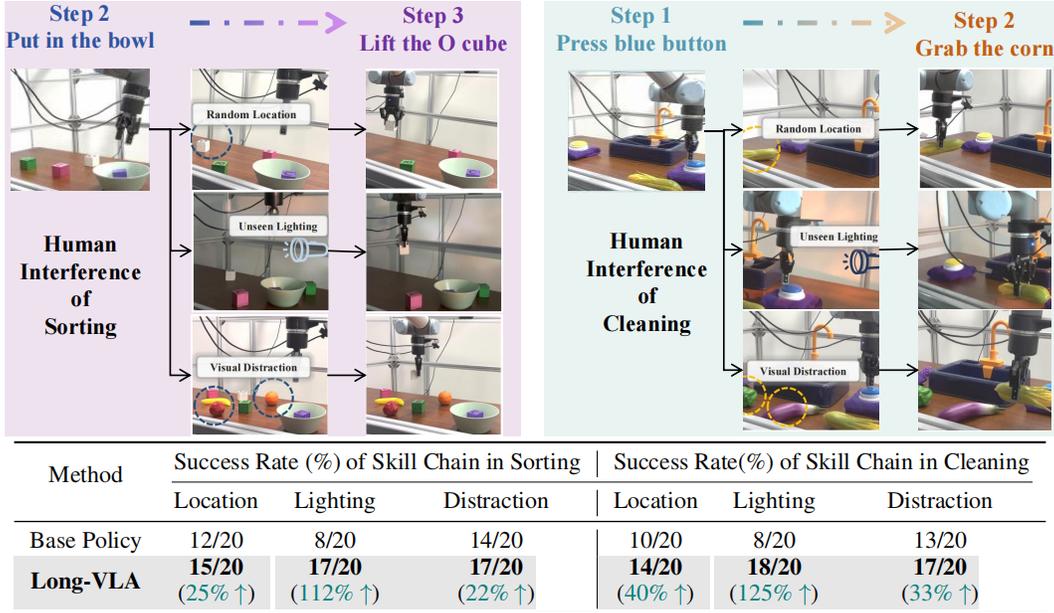

| Method | Success Rate (%) of Skill Chain in Sorting | | | Success Rate(%) of Skill Chain in Cleaning | | |
| --- | --- | --- | --- | --- | --- | --- |
| | Location | Lighting | Distraction | Location | Lighting | Distraction |
| Base Policy | 12/20 | 8/20 | 14/20 | 10/20 | 8/20 | 13/20 |
| **Long-VLA** | **15/20** (25% ↑) | **17/20** (112% ↑) | **17/20** (22% ↑) | **14/20** (40% ↑) | **18/20** (125% ↑) | **17/20** (33% ↑) |

Figure 10: Underlying Cause Analysis.

### D.3 Underlying Cause Analysis.

To further investigate the factors underlying the observed performance improvements, we conducted a validation experiment, as illustrated in Figure 10. For the sorting task, after completing step 2 (putting the cube into the bowl), we deliberately introduced perturbations to the position of the next target object, scene illumination, and added visual distractors before evaluating the subsequent task. Similarly, for the cleaning task, we applied comparable augmentations after completing step 1 (pressing the button). These additional interferences, introduced after the completion of each preceding task, result in a suboptimal initial state for the subsequent task. This leads to significant performance degradation in the base policy, with success rates dropping by approximately 50%. In contrast, our proposed approach consistently maintains a success rate of around 80%, demonstrating strong robustness during the skill chaining phase. This robustness is a key reason why our model's performance remains stable and does not experience significant drops in long-horizon tasks. Moreover, the performance improvement achieved by our method is generally more pronounced in the more challenging cleaning task compared to sorting, which aligns with the differences observed in Figure 5 and Figure 6.

### D.4 Ablation on Input Modality

Here, in Table 6, it is evident that using detection and third-person view in the Moving phase, and using detection and first-person view in the Interaction phase, constitutes the optimal configuration.

### D.5 Learable Masking Strategy.

There are concerns that phase-aware masking may be less effective in complex tasks such as obstacle avoidance. Nevertheless, its strong performance on most tabletop and relatively simple tasks highlights its effectiveness in addressing the skill-chaining problem. For phase-based masking, we draw on conclusions from prior works such as Plan-Seq-Learn. Furthermore, by making the masking learnable, we observe that the moving stage tends to activate third-person views, while the interaction stage activates first-person views. As shown in Table 7, the outcome consistent with the design of our method.

Table 6: Ablation on Input Modality on CALVIN(D-D). d denotes detection information, s denotes static camera views, g denotes gripper camera views.

| Setting | Moving | | | Interaction | | | Avg. Len ↑ |
|---|---|---|---|---|---|---|---|
| | d | s | g | d | s | g | |
| w Decomposition | ✓ | ✓ | | ✓ | | ✓ | 4.81 |
| w Input-level Adaptation | ✓ | ✓ | | ✓ | ✓ | ✓ | 4.13 |
| w Unified Training | ✓ | ✓ | ✓ | | | ✓ | 4.66 |
| | | ✓ | | | | ✓ | 3.65 |
| | ✓ | ✓ | ✓ | ✓ | ✓ | ✓ | 4.48 |

Table 7: Rate of learnable masking in different phases.

| Phase | Masking | | |
|---|---|---|---|
| | static masking | gripper masking | no masking |
| Moving | 4.37% | 86.71% | 8.91% |
| Interaction | 87.78% | 2.68% | 9.54% |

### D.6 More Experiments.

**More Comparison of Long-VLA and $\pi_0$.** Here, we provide additional comparisons under unseen lighting conditions and visual distractions in Figure 11. Our Long-VLA also outperforms $\pi_0$ in these more challenging, unseen scenarios.

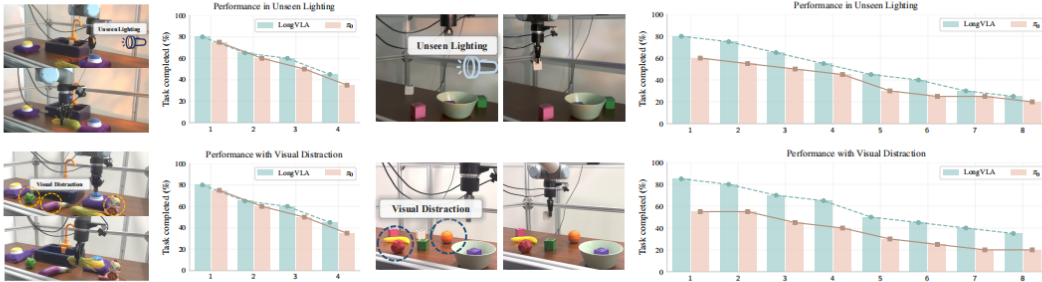

Figure 11: **More Comparison on Real World Scenarios (Left: Cleaning; Right: Sorting).**

**More Comparison of Simulation Environment.** Here, we also include additional comparisons on the ABCD→D split of the L-CALVIN simulation benchmark in Table 8.

**More Visualization in Real-world Scenarios.** Here, we offer more visualization in Figure 12, Figure 13 and Figure 14.

**More Visualization in Simulation Scenarios.** Here, we offer more visualization in Figure 15.

Table 8: Comparison with SOTA methods on L-CALVIN simulation benchmark.

| Train→Test | Method | Tasks Completed in Sequence | | | | | | | | | | Avg. Len |
|---|---|---|---|---|---|---|---|---|---|---|---|---|
| | | 1 | 2 | 3 | 4 | 5 | 6 | 7 | 8 | 9 | 10 | |
| ABCD→D | VLAS | 0.88 | 0.80 | 0.72 | 0.58 | 0.49 | 0.46 | 0.34 | 0.28 | 0.29 | 0.16 | 5.00 |
| | GR-1 | 0.92 | 0.81 | 0.71 | 0.63 | 0.58 | 0.54 | 0.50 | 0.40 | 0.30 | 0.29 | 5.68 |
| | RoboVLMs | 0.90 | 0.80 | 0.78 | 0.70 | 0.64 | 0.60 | 0.51 | 0.41 | 0.36 | 0.34 | 6.04 |
| | **Long-VLA** | **1.00** | **1.00** | **0.98** | **0.91** | **0.85** | **0.82** | **0.79** | **0.70** | **0.63** | **0.56** | **8.24** |

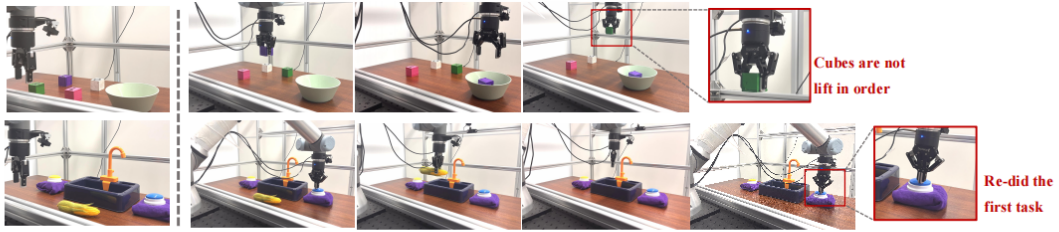

Figure 12: **Failure case of $\pi_0$.**

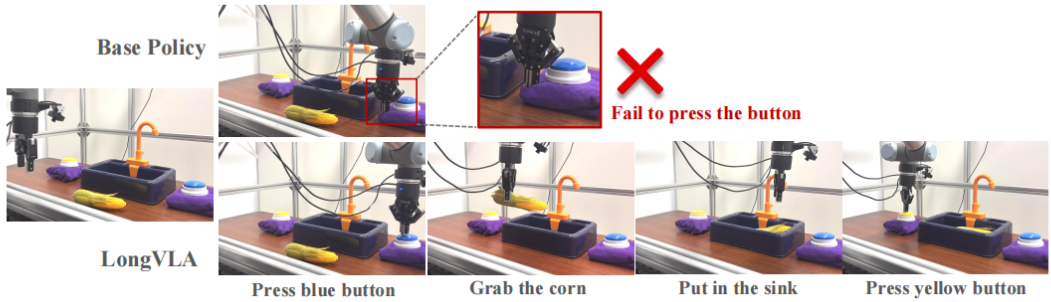

Figure 13: **Comparison of Execution in Sorting Task**

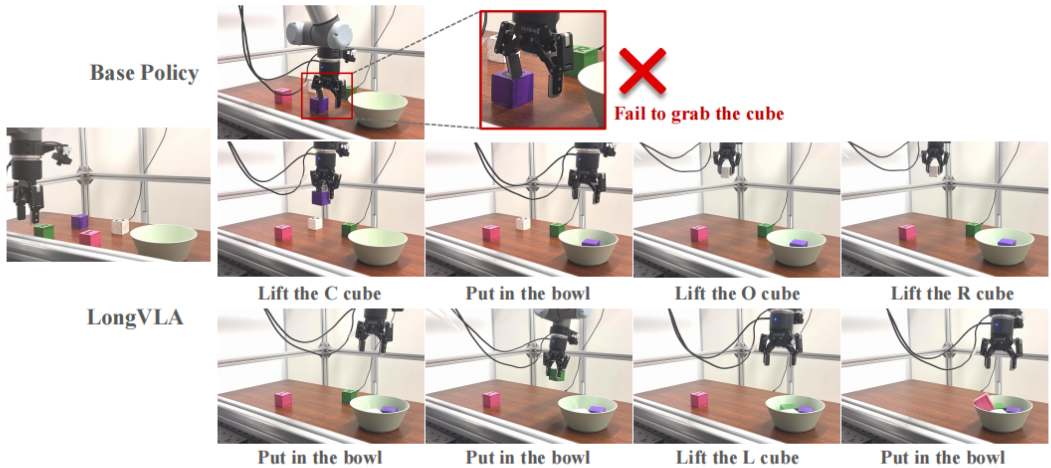

Figure 14: **Comparison of Execution in Cleaning Task**

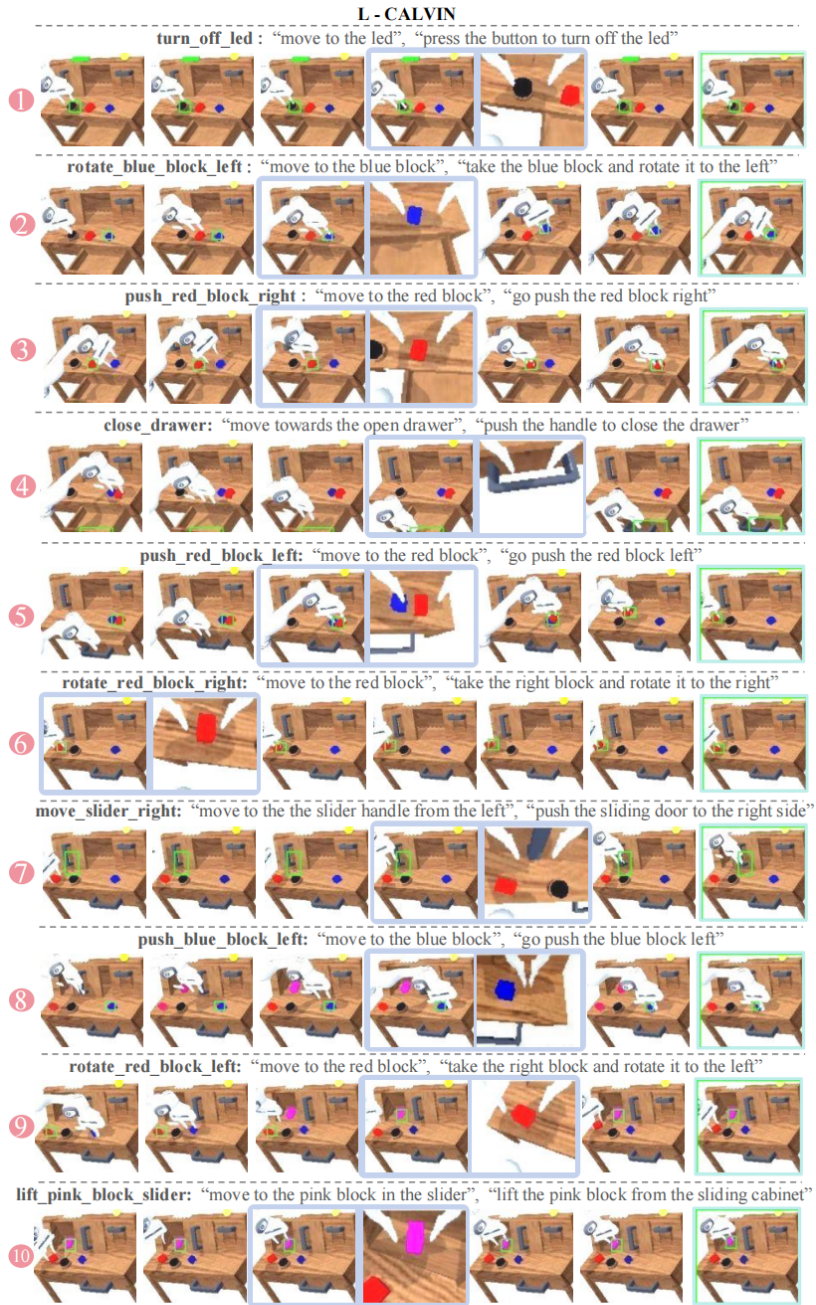

Figure 15: Rollouts of the L-CALVIN Benchmark. Visualization of Long-VLA rollouts on the D → D split of the 10-length L-CALVIN benchmark. The figure highlights gripper camera views during the switching moments between the moving policy and interaction policy under Long-VLA, demonstrating the importance of a good initial state for completing the task.

9